\documentclass[runningheads]{llncs}
\usepackage{graphicx}
\usepackage{import}
\usepackage{cite}
\usepackage{hyperref}

\usepackage{xcolor}  

\begin{document}
\title{Concept for a Technical Infrastructure for Management of Predictive Models in Industrial Applications}
\titlerunning{Infrastructure for Management of Predictive Models}\footnotetext[1]{The final publication is available at \url{https://link.springer.com/chapter/10.1007/978-3-030-45093-9_32}} 
%
\author{Florian Bachinger\orcidID{0000-0002-5146-1750} \and \\
Gabriel Kronberger\orcidID{0000-0002-3012-3189}}
\authorrunning{Bachinger et al.}
 
\institute{Josef Ressel Center for Symbolic Regression \\
Heuristic and Evolutionary Algorithms Laboratory  \\
University of Applied Sciences Upper Austria, Hagenberg, Austria
\email{florian.bachinger@fh-hagenberg.at}
}

\maketitle 

\begin{abstract}
  With the increasing number of created and deployed prediction models and the
  complexity of machine learning workflows we require so called model management
  systems to support data scientists in their tasks. In this work we describe
  our technological concept for such a model management system. This concept
  includes versioned storage of data, support for different machine learning
  algorithms, fine tuning of models, subsequent deployment of models and
  monitoring of model performance after deployment. We describe this concept
  with a close focus on model lifecycle requirements stemming from our industry
  application cases, but generalize key features that are relevant for all
  applications of machine learning. 
  \keywords{model management, machine learning workflow, model lifecycle, software architecture concept}
\end{abstract}
\section{Motivation}
  In recent years, applications of machine learning (ML) algorithms grew
  significantly, leading to an increasing number of created and deployed
  predictive models. The iterative and experimental nature of ML workflows
  further increases the number of models created for one  particular ML use
  case. We require so called model management systems to support the full ML
  workflow and to cover the whole lifecycle of a predictive model. Such a
  model management system should improve collaboration between data scientist,
  ensure the replicability of ML pipelines and therefore increase trust in the
  created predictive models. To highlight the need for model management
  systems we follow a typical machine learning process and identify shortcomings
  or problems occurring in practice, that could be mitigated by a model
  management system.

\subsection{Model management in the machine learning workflow}
\label{sec:typicalMLWorkflow}
\begin{figure}[h]
    \centering{
      \resizebox{0.6\linewidth}{!}{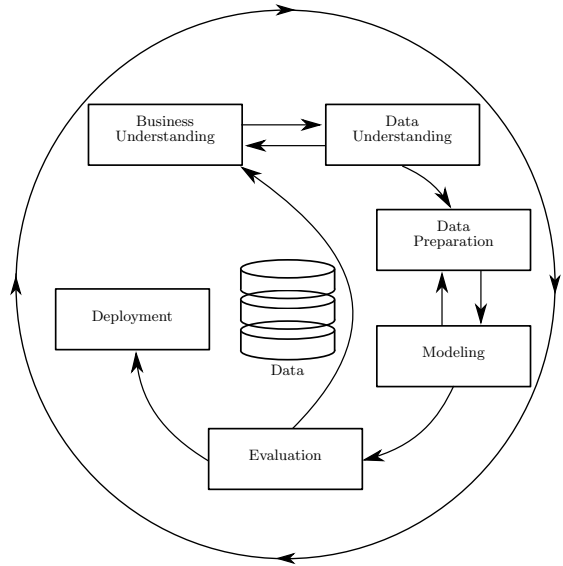}
      \caption{Visualization of the typical machine learning workflow as described in the CRISP-DM 1.0 Step-by-step data mining guide~\cite{Chapman2000}} 
      \label{fig:crisp_dm_process}
    }
  \end{figure}

  A typical ML workflow, as described by the CRISP-DM data mining
  guide~\cite{Chapman2000} and illustrated in Figure~\ref{fig:crisp_dm_process},
  is a highly iterative process. \textit{Business Understanding} and
  \textit{Data Understanding} is gained through assessment of the particular ML
  use case and the initial gathering and analysis of data. Meticulous
  \textit{Data Preparation}, cleaning of data and feature engineering, is an
  important prerequisite for good modeling results, as selection and data
  transformation are critical for successful application of ML methods. In the
  subsequent \textit{Modeling} task, different ML frameworks and algorithms are
  applied. Therein, it is necessary to test a variety of algorithm
  configurations. It is common that the data preparation step and modeling step
  are repeated and fiddled with, until satisfying results are achieved. Analysis
  of applied data preparation techniques and their affect on model quality can
  provide insights on the physical system and improve future modeling tasks in
  this domain. Similarly, comparison of all \textit{Evaluation} results can
  provide additional insights about suitable algorithm configurations for
  similar ML problems.

  Finally, when a suitable model is discovered, it is deployed to the target
  system. Typical ML workflows often end with this step. However, because of
  changes to the system's environment, i.e. concept drift, the model's
  predictive accuracy can deteriorate over time. Information about the several
  ML workflow steps that led to the deployed model might be forgotten, lost, or
  scattered around different files or knowledge systems. Model management
  systems should aid by tracking the complete ML workflow and saving every
  intermediate artefact in order to ensure replicability.

\subsection{Managing the model lifecycle - industrial applications}
  Predictive models are increasingly deployed to so-called edge computing
  devices, which are installed close to the physical systems, e.g. for
  controlling production machines in industry plants. In such scenarios,
  predictive models usually need to be tuned for each particular installation
  and environment, resulting in many different versions of one model.
  Additionally, models need to be updated or re-tuned to adapt to slowly
  changing systems or environmental conditions, i.e. concept drift. Once a tuned
  model is ready for deployment, the model needs to be validated to ensure the
  functional safety of the plant. The heterogenous landscape of ML frameworks,
  their different versions and software environments, further increases
  difficulty of deployment. We argue that model management system need to ML
  should borrow well established concepts from software development, i.e.
  continuous integration, continuous delivery, to cope with fast model
  iterations, and to cover the whole model lifecycle.

  In a subsequent phase, the deployed model's prediction performance, during
  production use, needs to be monitored to detect concept drift or problems in
  the physical system. Continuous data feedback from the physical system back to
  the model management system provides additional data for training, and future
  model validation.

\section{Related work}

  Kumar et al. \cite{Kumar2017} have fairly recently published a survey on
  research on data management for machine learning. Their survey covers
  different systems, techniques and open challenges in this areas. Each surveyed
  project is categorized into one of three data centric categorizations: 
  \begin{description}
    \item[ML in Data Systems] cover projects that combine ML frameworks with
    existing data systems. Projects like Vertica~\cite{Prasad2015},
    Atlas~\cite{Wang2003} or Glade~\cite{Cheng2012} integrate ML functionality
    into existing DBMS system by providing user-defined aggregates that allow
    the user to start ML algorithms in an SQL like syntax, and provide models as
    user-defined functions.
    \item[DB-Inspired ML Systems] describe projects that apply DB proven
    concepts to ML workloads. Most projects apply these techniques in order to
    speedup or improve ML workloads. This includes techniques like asynchronous
    execution, query rewrites and operator selection based on data clusters, or
    application of indices or compression. ML.NET Machine Learning as described
    by Interlandi et al.~\cite{Interlandi2018} introduces the so called DataView
    abstraction which adapts the idea of views, row cursors or columnar
    processing to improve learning performance.
    \item[ML Lifecycle Systems] go beyond simply improving performance or
    quality of existing ML algorithms. These systems assist the data-scientist
    in different phases of the ML workflow. In their survey, Kumar et al.
    \cite{Kumar2017} further detail the area of \textit{ML Lifecycle Systems}
    and introduce so called \textit{Model Selection and Management} systems.
    These systems assist not one but many phases of the ML lifecycle. 
  \end{description}

  One representative of \textit{Model Selection and Management} systems is
  described by Vartak et al.~\cite{Vartak2016}. Their so called ModelDB, is a
  model management system for the spark.ml and scikit-learn machine learning
  frameworks. ModelDB provides instrumented, wrapped APIs replacing the standard
  Python calls to spark.ml or scikit-learn. The wrapped method calls the ML
  framework functionality and sends parameters or metadata of the modelling
  process to the ModelDB-Server. This separation allows ModelDB to be ML
  framework agnostic, given that the ModelDB API is implemented. Their system
  provides a graphical user interface (GUI) that compares metrics of different
  model versions and visualizes the ML pipeline which lead to each model as a
  graph. However, ModelDB only stores the pipeline comprised of ML instructions
  that yielded the model. ModelDB does not store the model itself, training
  data, or metadata about the ML framework version. External changes to the
  data, for example, are not recognized by the system and could hamper
  replicability.

  Another representative, ProvDB, as described by Miao et al.~\cite{Miao2017}
  uses a version control system (e.g. git) to store the data and script files
  and model files created during the ML lifecycle in a versioned manner.
  Therein, scripts can be used for either preprocessing or to call ML framework
  functionalities. Git itself only recognizes changes to the files and therefore
  treats changes to data, script or model files equally. In order to store
  semantic connections between e.g. data versions and their respective
  preprocessing scripts or the connection between data, the ML script and the
  resulting model, ProvDB uses a graph database (e.g. Neo4j) on top. Provenance
  of files and metadata about the ML workflow, stored in the graph database, can
  be recorded through the ProvDB command line or manually defined through the
  file importer tool or the ProvDB GUI. This design allows ProvDB to support
  data and model storage of any ML framework given that these artefacts can be
  stored as files and are committed to git. However, automatic parsing and
  logging of instructions to the ML framework only works in the ProvDB command
  line environment. Inside the ProvDB environment, calls to the ML framework and
  their parameters are first parsed and then forwarded which requires ML
  frameworks that provide a command line interface.
  
  Similar to ProvDB we aim to store all artefacts created during the ML
  lifecycle and allow the definition of semantic connections between these
  artefacts. Our approach differs from ProvDB as we plan to use a relational
  database for the data persistence and aim to develop a tighter integration to
  the actual ML framework, comparable to the API approach of ModelDB. Moreover,
  we aim to support tuning, automated validation and deployment of ML models and
  provide functionality to monitor performance of deployed models.

  In their conclusion Kumar et al. \cite{Kumar2017} identify the area of
  "\textit{Seamless Feature Engineering and Model Selection}", systems that
  support end-to-end ML workflows, as important open areas in the field of data
  management for machine learning. They highlight the need for fully integrated
  systems that support the machine learning lifecycle, even if it only covers a
  single ML system/framework. 

\section{Design and architecture}
\begin{figure}[h]
  \centering{
    \resizebox{\linewidth}{!}{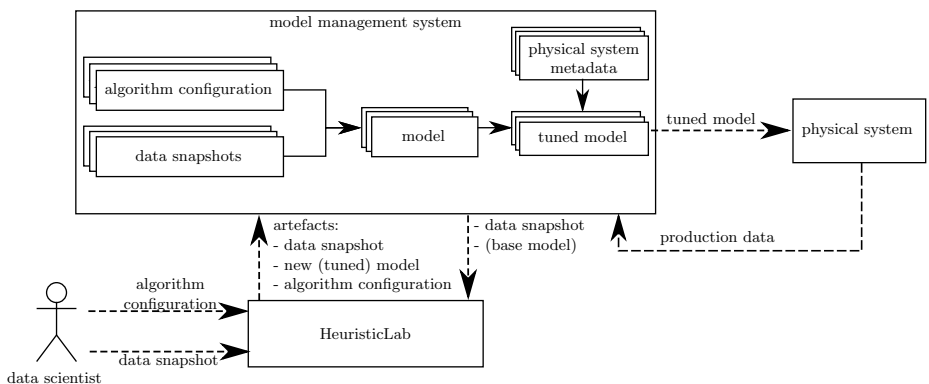}
    \caption{Visualization of interaction between the envisioned model management system, a machine learning framework (e.g. HeuristicLab) and an external physical target system.} 
    \label{fig:architectureIdea}
  }
\end{figure}

In the following section we describe our concept of a model management system.
This system is designed to be ML framework agnostic. Integration into the open
source ML framework HeuristicLab\footnote{\url{https://dev.heuristiclab.com} },
which is being developed and maintained by our research group, will serve as a
prototypical template implementation. Figure~\ref{fig:architectureIdea} serves
as illustration of the data-flows and individual components of the model
management system described in the following sections. The described system can
be used either locally by a single user or as a centralized instance to enables
collaboration of different users. The model management system is described by
the following key features:
\begin{itemize}
  \item Centralized and versioned data storage for all artefacts of the ML
  framework.
  \item Definition of semantic connections between the different artefacts.
  \item Storage API for ML framework integration.
  \item Automatic evaluation of models on semantically connected snapshots.
  \item Bundling of models for deployment and subsequent monitoring.
\end{itemize}

\vspace{-0.5cm}
\subsection{Data management}
  The accuracy of prediction models, achievable by different ML algorithms,
  depends on the quality of the (training) data. Errors in data recording, or
  wrong assumptions made during business- and data-understanding phase affect
  data preparation and are therefore carried over to the modeling phase and will
  subsequently result in bad models. Though, bad models are not solely caused by
  poor data quality, as a model can become biased if certain information,
  contained in the dataset, was then not present in the training portion of the
  ML algorithm. These problems are especially hard to combat or debug if the
  connection between a specific model version and its training data was not
  properly documented or if the information is scattered around different
  knowledge bases and therefore hard to connect and retrieve. A model management
  system should therefore provide an integrated, versioned data-storage. 

  ML frameworks and supported preprocessing tools need to be able to query
  specific version of data, i.e. snapshots, from the database. Preprocessing
  tools also need to be able to store modified data as new snapshots.
  Additionally, data scientists should be able define semantical relations
  between snapshots. This connections can be used to mark compatible datasets
  that stem from similar physical systems or are a more recent data recording of
  the same system, as also discussed in Section~\ref{sec:dataFeedback}. In such
  cases a model management system could automatically evaluate a model's
  prediction accuracy on compatible snapshots. The same semantic connections can
  be used to connect base datasets with the specific datasets from physical
  systems for model tuning. When a new, better model on the base dataset is
  created the model management system can automatically tune it to all connected
  specific datasets.

\subsection{Model management}
  The section model management loosely encompasses all tasks and system
  components related to the ML model, this includes the ML training phase and
  the resulting predictive model, fine tuning the model to fit system specific
  data sets, evaluation of models (on training sets or physical system data) and
  the subsequent deployment of validated models.

  \subsubsection{Model creation or model training}
  In order to conveniently support the ML workflow, a model management system
  should impose no usability overhead. In case of the \textit{Modeling} phase
  this means that necessary instrumentation of ML framework methods, to capture
  ML artefacts, should not affect usage or require changes in existing
  pipelines/scripts. Therefore, method signatures of the ML framework must stay
  the same. Functionally, the model management system has to be able to capture
  all ML framework artefacts, configurations and metadata necessary to fully
  reproduce the training step. In our concept for a model management system we
  intend to provide an easy to use API for capturing artefacts that can be
  integrated by any ML framework. The resulting knowledge base of tried and
  tested ML algorithm parameters for a variety of ML problems can serve as a
  suggestion for suitable configurations for future experiments, or meta-heuristic optimization for problem domains.

  \subsubsection{Model evaluation}
  Besides ensuring replicability of the ML workflow, a model management system
  should also aid in the evaluation of models. In practice we require evaluation
  of a model's prediction accuracy not only on the test section of data, but
  also on "older" data snapshots, to evaluate whether a new model actually has
  achieved equal or better predictive quality than it's predecessor. 

  Similarly, functional safety of the prediction models in their production
  environments can be ensured by automated validation of models on past
  production data or on simulations of the physical system. This model
  evaluation process can be seen as the analogy of unit tests in the continuous
  development process. Facilitation of the semantic connections between datasets
  provides the necessary information these evaluation steps.

  \subsubsection{Model tuning}
  Predictive models often need to be tuned in order to describe a specific
  target system. Model tuning refers to the task of using ML algorithms to adapt
  an existing predictive model, or model structure, to fit to a specific
  previously unknown environment. Model tuning can be used to fit an existing
  model to its changed environment after a concept drift was detected. Likewise,
  tuning can improve or speedup the ML workflow by using an existing, proven
  model as a starting point to describe another representative of a similar
  physical system. If a model type and ML framework support tuning,  the model
  management system can trigger this tuning process and subsequent evaluation
  automatically. This process reassembles automated software build processes. If
  concept drift is detected or automated model tuning is enabled for a physical
  system, the model management system can take action  autonomously.

  \subsection{Model deployment}
  The model management system should aid in the deployment of prediction models.
  This means providing the model bundled with all libraries necessary for
  execution of the model. The heterogenous landscape of ML frameworks and their
  different versions and software environments can cause compatibility issues on
  the target system. Crankshaw et al.~\cite{Crankshaw2017} describe a system
  called Clipper, that solves this problem by deploying the model and its
  libraries bundled inside Docker images. This technique could also aid in
  distribution of the model, as the Docker ecosystem includes image management
  applications, that host image versions and provide deployment mechanisms. By
  applying this technique, the model management system can ensure executability
  and solve delivery of models to the target system. 

  \subsection{Data feedback}
  \label{sec:dataFeedback}
  Our concept for a model management system includes a data gathering component
  to capture feedback in the form of production data from the edge device.
  Monitoring of the model's prediction accuracy during deployment and evaluation
  of the model on the production data enables the model management system to
  detect concept drift and to tune and subsequently re-deploy tuned models. The
  software necessary for monitoring and data gathering can be deployed to the
  edge device by addition to the bundle created during model deployment.

\section{Summary}
In this work we described our technological concept for a model management
system. We describe the different features and components that are necessary to
fully capture the machine learning lifecycle to ensure replicability of modeling
results. Application of predictive models in industrial scenarios provides
additional challenges regarding validation, monitoring, tuning and deployment of
models that are addressed by the model management system. We argue that advances
in model management are necessary to facilitate the transition of machine
learning from an expert domain into a widely adopted technology. 

\subsection*{Acknowledgement}
The financial support by the Christian Doppler Research Association, the
Austrian Federal Ministry for Digital and Economic Affairs and the National
Foundation for Research, Technology and Development is gratefully acknowledged.

\bibliographystyle{splncs04}
\bibliography{references}

\end{document}